\documentclass{article}

\usepackage{arxiv}

\usepackage[utf8]{inputenc} %
\usepackage[T1]{fontenc}    %
\usepackage{hyperref}       %
\usepackage{url}            %
\usepackage{booktabs}       %
\usepackage{amsfonts}       %
\usepackage{nicefrac}       %
\usepackage{microtype}      %
\usepackage{lipsum}         %
\usepackage{graphicx}
\usepackage{biblatex}
\usepackage{doi}

\usepackage{amsmath}
\usepackage{cleveref}       %
 \usepackage[labelfont=bf]{caption}

\usepackage{color}
\usepackage{wrapfig}
\usepackage{times}
\usepackage{caption}
\newcommand{\methodname}[1]{MyoSuite}

\title{MyoSuite \\  A contact-rich simulation suite for musculoskeletal motor control}

\author{
  Vittorio Caggiano* \\
  Meta AI Research, New York, NY, USA \\
  \texttt{caggiano@gmail.com} \\
  \And
  Huawei Wang* \\
  University of Twente, the Netherlands \\
  \texttt{h.wang-2@utwente.nl} \\
  \And
  Guillaume Durandau \\
  University of Twente, the Netherlands \\
  \texttt{g.v.durandau@utwente.nl} \\
  \And
  Massimo Sartori \\
  University of Twente, the Netherlands \\
  \texttt{m.sartori@utwente.nl} \\
  \And
  Vikash Kumar \\
  Meta AI Research, Pittsburgh, PA, USA \\
  \texttt{vikashplus@gmail.com} \\
}

\begin{document}

\maketitle

\begin{abstract}%

Embodied agents in continuous control domains have been traditionally exposed to tasks with limited opportunity to explore musculoskeletal details that enable agile and nimble behaviors in biological beings. The sophistication behind bio-musculoskeletal control not only poses new challenges for the learning community but realizing agents embedded in the same perception-action loop that the human sensory-motor system solves can also have a far-reaching impact in fields of neuro-motor disorders, rehabilitation, assistive technologies, as well as collaborative-robotics.

Human biomechanics is a complex multi-joint-multi-actuator musculoskeletal system. The sensory-motor system relies on a range of sensory-contact rich and proprioceptive inputs that define and condition motor actuation required to exhibit intelligent behaviors in the physical world. Current frameworks for studying musculoskeletal control do not include at the same time the needed physiological sophistication of the musculoskeletal systems and support physical world interaction capabilities. In addition, they are neither embedded in complex and skillful motor tasks nor are computationally effective and scalable to study motor learning in the timescale that current learning paradigms require.

To realize a platform where physiological detail and challenges behind human motor control can be investigated, we present MyoSuite -- a suite of physiologically accurate biomechanical models of elbow, wrist, and hand, with physical contact capabilities which allow complex and skillful contact-rich real-world tasks. The implemented motor tasks provide a great variability of control challenges: from simple postural control to skilled hand-object interactions involving tasks like turning a key, twirling a pen, rotating two balls in one hand, etc. Finally, by supporting physiological alterations in musculoskeletal geometry (tendon transfer), assistive devices (exoskeleton assistance), and muscle contraction dynamics (muscle fatigue, sarcopenia), we present real-life tasks with temporal changes, thereby exposing realistic non-stationary conditions in our tasks which most continuous control benchmarks lack. \\
Project Webpage: \href{https://sites.google.com/view/myosuite}{\color{blue}https://sites.google.com/view/myosuite} \\
Github: \href{https://github.com/facebookresearch/myosuite}{\color{blue}https://github.com/facebookresearch/myosuite}

\end{abstract}

\section{Introduction}

\begin{figure}[h!]
\centering
\includegraphics[width=\textwidth]{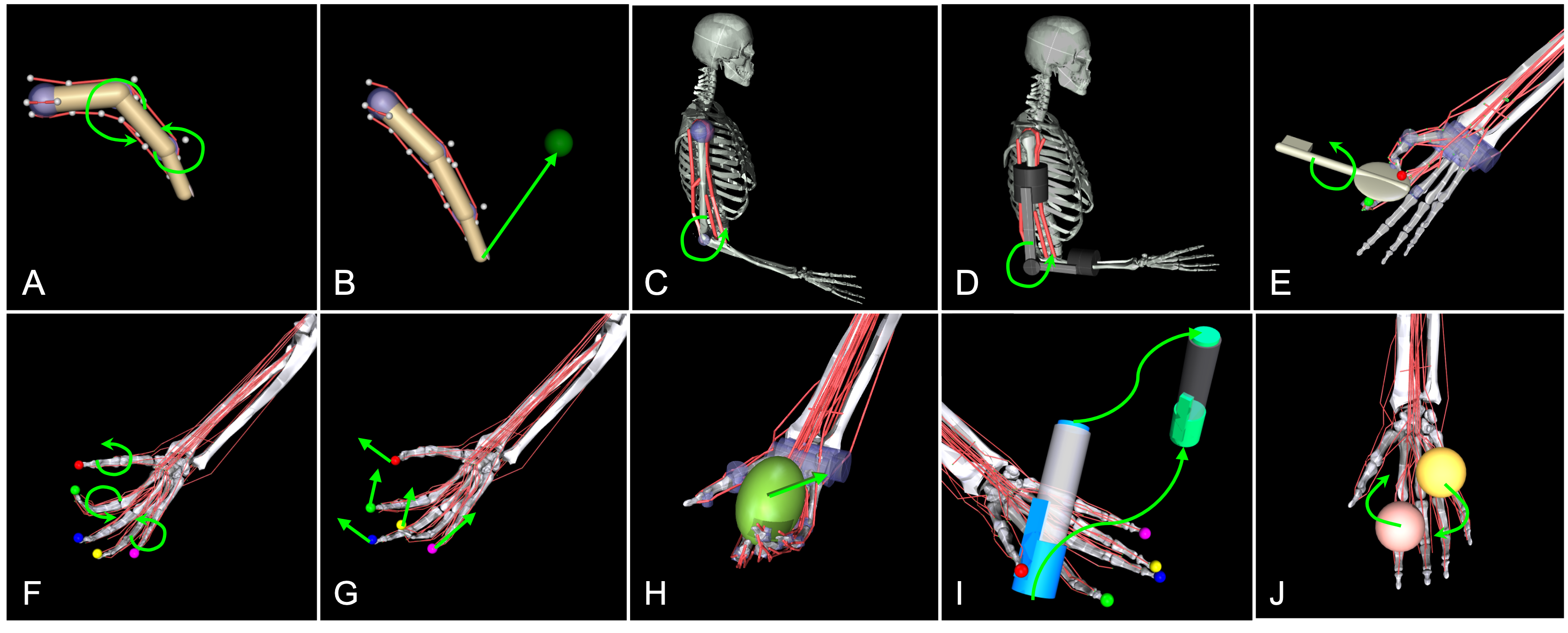}
  \caption{MyoSuite tasks -- \textbf{Finger}: joint pose (A), tip reach (B). \textbf{Elbow}: joint pose (C), exo-assisted (D). \textbf{Hand}: key turn (E), joint pose (F), tips reach (G),  object reposition (H), pen-twirl  (I),  baoding balls (J).}
 \label{Fig:DexterousTasks}
\end{figure}

Data-driven learning paradigms have enabled rapid progress in multiple fields such as vision (\cite{Deng_2009}), natural language processing (\cite{GLUE}), speech (\cite{Librispeech}), among others. While there has been progress, the field of embodied AI is still awaiting its breakthrough moments. As opposed to fields in which decision making is one step open loop, embodied agents operate decisions in a close action-perception loop over time, which adds complexities to the problem and the solutions. Traditionally this has been approached exposing agents to simplified problems, that, although have pushed the field forward do not translate into real-world advancements. One of the root causes is that real-world problems are complex and non-stationary. The human central nervous system, on the other side, can easily handle non-stationary and complex tasks (\cite{latash2012bliss}) by building control schemes that integrate and model proprioceptive and sensory information and transform them into optimal (\cite{Shadmehr2008, Scott2012}) and adaptable  (\cite{Wolpert2000ComputationalPO}) control policies. Such neuro-control challenges that the human central nervous system can seamlessly handle can pose as the next frontier for continuous control algorithmic paradigms. Furthermore, any development in such a critical problem space can translate into meaningful advancements in critical fields like neuromechanics, physiotherapy, rehabilitation, as well as robotics.

The musculoskeletal system presents more complex and challenging problems for embodied agents as opposed to virtual and robotic systems. Muscles can only generate forces in one direction (pulling, not pushing); and undergo changes in their force-generating capacity depending on their operating length, contractile velocity, or fatigue state (\cite{Arnold2011FibreOL}). Moreover, muscles may undergo structural changes in response to aging e.g. sarcopenia, or to exercise, thereby further altering their contractile properties across time (\cite{Age_Exc_Muscle_1998}). Surgical interventions and assistive devices e.g. exoskeleton can also alter the muscle actuation. In addition, the muscle control space is also high dimensional (the number of muscles exceeds the number of human joints - about 600 muscles to control about 300 joints), redundant (multiple muscles act on the same joint), and multi-articular (muscles very often act on multiple joints) (\cite{Hirashima_2016,Ting_2012,DeGroote2016EvaluationOD}). As a result, the overall system suffers from the `curse of dimensionality' (\cite{bernstein1966co}) as well as non-stationarity challenges, which are less common in joint level control, which is typical in robotics (\cite{Wolpert2000ComputationalPO}).

The machine learning community has made major advancements in the embodied AI field by defining benchmarks and baselines. OpenAI-Gym (\cite{brockman2016openai}) and DmControl (\cite{tassa2018deepmind}) are the \textit{de-facto} benchmarks for behavior synthesis in continuous control domains. These benchmarks however are not suitable for real-world problems. First, they consists mostly of simple problems\footnote{With some exceptions like \textit{HandManipulateEgg} and \textit{HandManipulatePen} (\cite{Multi-goalRL}) or in-hand manipulation of real-world objects (\cite{openai2019solving, huang2021generalization,chen2021general})} and are largely already solved. Second, those benchmarks have very limited capabilities to test the adaptability of an agent in response to non-stationary environment changes. Usually, those environment changes have been investigated via contrived examples - removing links (\cite{nagabandi2018learning}) or adding spurious joints (\cite{gupta2017learning, devin2017learning}). There is a dire need for new benchmarks embedded closely in the real world which, at the same time, create new challenges for algorithmic paradigms and translate into real-world impact.

\begin{wraptable}{r}{8.2cm}
    \footnotesize

    \begin{center}
    \begin{tabular}{| c c c c|}
    \hline
    & Contact & Muscle & Speed    \\
    & Dynamics & Model &     \\
    \hline    MuJoCo(MyoSuite) & $\textbf{\textcolor{blue}{\checkmark}}$ & $\textcolor{blue}{\checkmark}$ & $\textcolor{blue}{\uparrow}$  \\
    OpenSim &  & {\checkmark} & $\downarrow$   \\
    AnyBody &  & \checkmark & $\downarrow$   \\
    SIMM &  & \checkmark & $\downarrow$   \\
    PyBullet & \checkmark &  & $\uparrow$  \\
    Dart & \checkmark & \ref{physc_msc_footnote} & $\uparrow$ \\
    RaiSim & \checkmark & \ref{physc_msc_footnote} & $\uparrow$ \\
    \hline
    \end{tabular}
    \end{center}

    \caption{Physic simulators. \methodname~ provides physiologically realistic and computationally efficient musculoskeletal models in the MuJoCo physics engine.}
    \label{Table:Simulators}

\end{wraptable}

Previous attempts at establishing realistic musculoskeletal tasks as benchmarks (\cite{Song2020.08.11.246801}) were narrowly defined. They favored bigger and functionally relevant muscle groups e.g. legs (\cite{hamner2010muscle, sartori2013musculoskeletal, white1989three, ackermann2010optimality}) and arms (\cite{OpenSim_Delt2007, OpenSim_Seth2018, Saul_2015, Mcfarlandl_2019, Lea_Asakawa_2015, saul2015benchmarking}). These attempts relied on physics-based musculoskeltal simulation engines such as \hyperlink{https://simtk.org/projects/opensim}{OpenSim} (\cite{OpenSim_Seth2018}), \hyperlink{https://www.anybodytech.com/software/ams/}{AnyBody} (\cite{damsgaard2006analysis}) and \cite{SIMM} that although are widely used for simulating human neural-mechanical control, human robot interaction, and rehabilitation are computationally expensive (simulating large number of muscle is intractable) and provide limited support for contact rich interactions with their environments (see Table \ref{Table:Simulators}). While, physics engines used in the robotic field (PyBullet \cite{coumans2016pybullet}, MuJoCo \footnote{Contains a basic implementation of a muscle model that \methodname~ builds upon and it has been used to develop musculoskeletal models of animals e.g. \cite{LaBarbera2021}} \cite{todorov2012mujoco}, IsaacGym \cite{IsaacGym_2021}, RaiSim \cite{raisim2018}, and Dart \cite{lee2018dart})\footnote{\label{physc_msc_footnote}Efforts have been made to add muscle models in both DART \cite{Seunghwan2019} and RaiSim \cite{Younguk2018}} are relatively more efficient and support contact interactions, but lack adequate support for modeling anatomical and functionally validated musculoskeletal models (see Table \ref{Table:Simulators}).

In order to expose the community to the exciting challenges presented by the musculoskeletal control of biological systems, we present a physiologically realistic and computationally efficient framework:  \textbf{\methodname}.

\paragraph{Our contributions:}
    \begin{itemize}

        \item A fully automated model-agnostic pipeline which reduce the tedious manual effort often required in creating new models by optimizing existing and validated (OpenSim) musculoskeletal models. The resulting models are numerically equivalent to the original and can be simulated two orders of magnitude faster than the original models

        \item We developed a set of musculoskeletal models (from simple one joint to complex full hand) in MuJoCo that are physiologically accurate, and supports full contact dynamics.

        \item We designed a family of 9 realistic dexterous manipulation tasks (from simple posing to simultaneous manipulation of two Baoding balls) using these models. These task families take inspiration from the state of art robotic dexterous manipulation results (\cite{DAPG, PDDM, OpenAI}).

        \item \methodname~ supports physiological alterations in musculoskeletal geometry (tendon transfer,  exoskeleton assistance) and muscle contraction dynamics (muscle fatigue,  sarcopenia) to expose real-life tasks with temporal changes, thereby subjecting self-adapting algorithms to the realistic non-stationary challenges under continuous control settings.

        \item In its present form \methodname~ consists of 204 tasks:  9 task-families x 2 difficulty level (easy, hard) x 3 reset conditions (fixed, random, and none\footnote{To facilitate investigation in reset-free algorithms such as \cite{MTRF}}) x 8 (or 4) combinations of non-stationarity variations. We present some baseline results on \methodname~ outlining its features and complexities.
    \end{itemize}

\section{Preliminaries}
Behavior synthesis for embodied agents can be formulated as a sequential decision-making problem where the goal is to generate a coordinated sequence of trajectories that allows the agent to achieve desired outcomes. We first provide a brief overview of the musculoskeletal system before outlining how its behavior synthesis can be formulated using Markov Decision Processes (MDP) formulations under the Reinforcement Learning (RL) paradigm.

\subsection{Musculoskeletal models} \label{NMSDetails}
Musculoskeletal systems consist of bones of various lengths connected together via a redundant system of skeletal muscles and tendons. Unlike robotic systems, which can be viably investigated by directly controlling and analyzing the resulting samples (e.g. parameter identification is solved comprehensively), mathematical models for biological systems are based, however, on sparse data available from cadavers (e.g. not all parameters can be identified leading to inherent uncertainty). Owing to the limitations of live experimentation, musculoskeletal models have been the cornerstone of most investigation and understanding behind biological motor control. Musculoskeletal models are commonly modeled as a 3rd order system which contains first or second-order muscle activation and contractile dynamics as well as the second-order body dynamics.

\begin{figure}[h!]
\centering
\includegraphics[width=.8\textwidth]{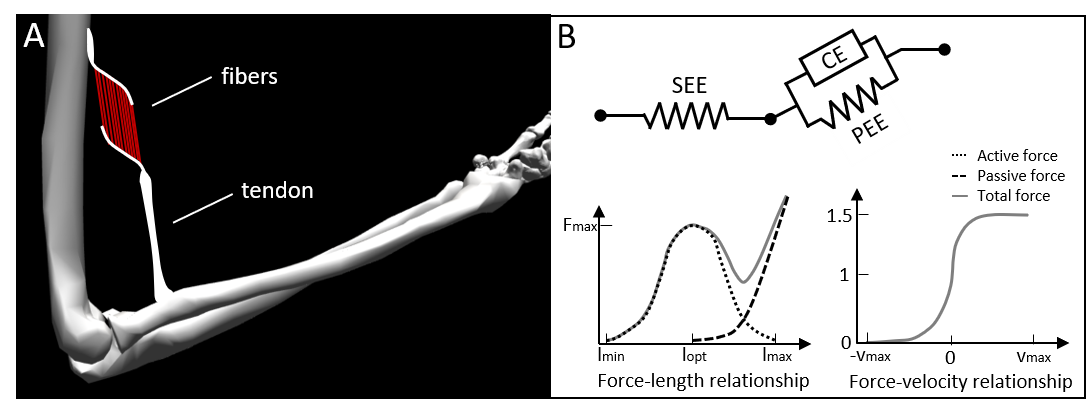}
  \caption{Example of muscle structure (A) and the Hill-Type model (B).}
 \label{Fig:muscleExample}
\end{figure}

A typical muscle-tendon unit consists of contractile fibers and series-elastic tendons (see Figure \ref{Fig:muscleExample}A). Contractile fibers are the active actuating elements of a muscle-tendon unit that are controlled by the signals generated from alpha motor neurons ($\alpha$ -MNs) in the spinal cord. A widely used mathematical model for this is the Hill-type model (see Figure \ref{Fig:muscleExample}B). It typically contains three elements: a contractile element (CE, Figure \ref{Fig:muscleExample}B), a parallel elastic component (fiber stiffness, PEE, Figure \ref{Fig:muscleExample}B), and a series elastic component (tendon stiffness, SEE, Figure \ref{Fig:muscleExample}B). Muscle forces are functions of both the muscle lengths and velocities, due to the force-length and force-velocity properties (see Figure \ref{Fig:muscleExample}B, Equation \ref{equ: 1}a,b). In MuJoCo, tendons were assumed infinitely stiff. A first order dynamics is normally used to represent the muscle twitch response (e.g. activation and and deactivation dynamics, see Equation \ref{equ: 1}c). Muscle-tendon units project mechanical forces onto skeletal bones. Geometrically, muscle-tendon units normally originates and insert with attachments points spanning different segments/bones, thereby contributing to move the joints in between by pulling segments/bones closer to each other (see Figure \ref{Fig:muscleExample}A). Dynamics of the skeleton system follows the Newton's laws and it is normally written in the format of dynamic equations (see Equation \ref{equ: 1}d). All together, a complete musculoskeletal model can be defined as a 3rd order system.

\begin{subequations}
    \label{equ: 1}
    \begin{align}
        F_{mtu} = F_m \cos\alpha \\
        \label{1a}
        F_m = a \cdot f_a(l) \cdot f_v(\dot{l}) + f_p(l) \\
        \label{1a}
        \dot{a} = (u - a) / \tau(u, a) \\
        \label{1a}
        M \cdot \ddot{q} + C(q, \dot{q}) \cdot \dot{q} + G(q) = F_{mtu} \cdot r
    \end{align}
\end{subequations}
where, $F_{mtu}$ represents the muscle-tendon unit force; $\alpha$ represents the pennation angle between tendon and muscle fibers; $F_m$ represents the muscle force; $a$ represents the muscle activation; $f_a$ is the active muscle force-length function; $f_v$ is the muscle force-velocity function; $f_p$ is the passive muscle force-length function; $l$ represents the muscle length; $\dot{l}$ represents the muscle velocity; $u$ represents the neural control signal; $\tau(u, a)$ is the time constant of the muscle activation and the deactivation dynamics; $M$ is the Mass/inertia matrix of the skeleton system; $q, \dot{q}, \ddot{q}$ are the joint angles, velocities, and accelerations; $C(q, \dot{q})$ is the Coriolis force term; $G(q)$ is the gravity force term; $r$ represents the moment arm of the muscle at the contributing joint.

\subsection{Markov decision process (MDP) formulation}
In RL paradigms, we operate in a Markov decision process $\mathcal{M} = (\mathcal{S}, \mathcal{A}, \mathcal{T}, \mathcal{R}, \rho, \gamma)$. Per usual notation, $\mathcal{S} \subseteq \mathbb{R}^n$ and $\mathcal{A} \subseteq \mathbb{R}^m$ represent the continuous state and action spaces respectively. The unknown transition dynamics is described by $s' \sim \mathcal{T}(\cdot|s,a)$. $\mathcal{R}: \mathcal{S} \rightarrow [0, R_{\max}]$ , $\gamma \in [0, 1)$, and $\rho$ represents the reward function, discount factor, and initial state distribution respectively. Policy is a mapping from states to a probability distribution over actions, i.e. ${\pi} : \mathcal{S} \rightarrow P(\mathcal{A})$, which is parameterized by $\theta$. The goal of the agent is to learn a policy $\pi^*_{\theta}(a|s) = argmax_{\theta}[J(\pi, \mathcal{M})]$, where $J = \max_{\theta} \mathbb{E}_{s_0 \sim \rho(s), a \sim \pi_{\theta}(a_t|s_t)}[\sum_t R(s_t, a_t)]$ i.e. the expected sum of discounted rewards in an episodic setting. Policy gradient algorithms (\cite{silver2014deterministic}), TD-learning based algorithms, such as Q-learning (\cite{watkins1992q}), SARSA (\cite{sutton1996generalization}), actor-critic based methods (\cite{konda2000actor}), etc. can be leveraged to optimize $J$ to generate behaviors.

\subsection{MDP formulation for Robotic systems vs. Musculoskeletal systems}

\textbf{Robotic systems}:
Under our MDP characterization $\mathcal{M}$, state space $\mathcal{S}$ consists of $\lbrace joint~position, \\  joint~velocity\rbrace$, action space $\mathcal{A}$ consists of actuator's $\lbrace position/velocity/torque~demands \rbrace$ and samples for policy optimization, are gathered either directly from the real world transition $\mathcal{T}_{real}$, or via physics simulation engines $\mathcal{T}_{sim}$ (\cite{todorov2012mujoco, coumans2016pybullet, lee2018dart}).

\textbf{Musculoskeltal systems}:
Under the MDP characterization $\mathcal{M}$, state space $\mathcal{S}$ consists of $\lbrace muscle-tendon~length,~muscle-tendon~velocity,$  $muscle~activations\rbrace$, action space $\mathcal{A}$ consists of actuator's $\lbrace \alpha -Motoneurons~signals \rbrace$\footnote{Motorneurons are the final neuronal stage that connects the central nervous system to the muscles.} samples for policy optimization, are gathered from the physics simulation engines $\mathcal{T}_{sim}$ controlled via muscle actuators.

\section{MyoSuite}
\begin{figure}[h!]
\centering
\includegraphics[width=\textwidth]{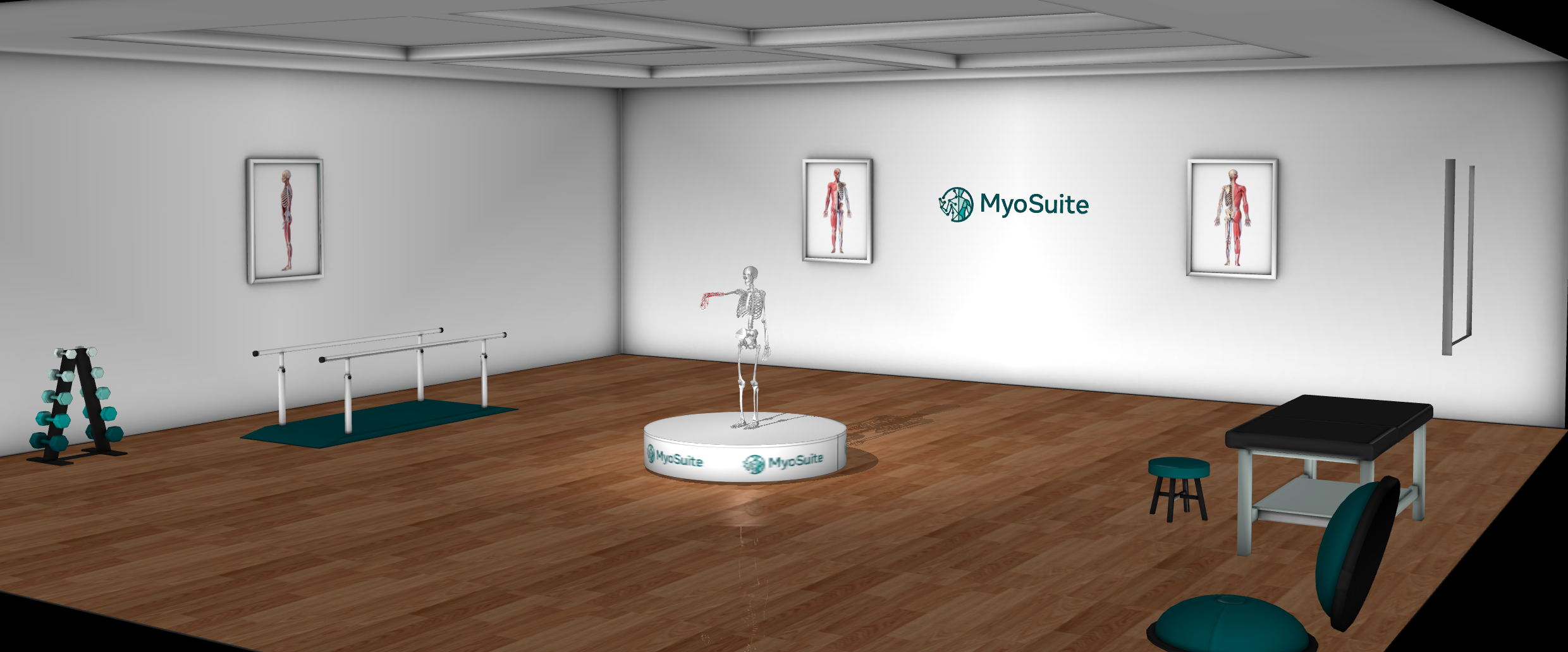}
  \caption{MyoSuite offers a playground to test several physiologically accurate musculoskeletal models challenged to perform skillful and contact-rich tasks.}
 \label{fig:MyoSuite}
\end{figure}

Unlike robotic counterparts, which are double acting and usually have one-to-one relationships between joints and actuators, tendons and muscles in musculoskeletal systems act via contraction (pull-only) and span multiple joints inducing strong coupling between them. Muscles become fatigued with extended usage. Tendons transfer muscle forces to bones while serving as temporary energy storage units for motion efficiency. These details, while complex, conceal within themselves the ingredients of effective motor control in biological systems. To facilitate investigations in these details, we present the \textit{``MyoSuite''} (see Figure \ref{fig:MyoSuite}) which consists of a collection of physiologically accurate musculoskeletal models (sec \ref{sec:models}) and physically realistic contact-rich tasks (sec \ref{sec:tasks}) of varying complexity.

\subsection{Models}
\label{sec:models}
Our models are developed using a thorough investigation of well studied existing models (\cite{OpenSim_Delt2007, OpenSim_Seth2018, Saul_2015, Mcfarlandl_2019, Lea_Asakawa_2015, saul2015benchmarking}) and functional studies (\cite{wu2008analysis}). We started from OpenSim models (\cite{Lea_Asakawa_2015, Mcfarlandl_2019, Saul_2015}) of the arm and hand which are widely used in fields of human neural-mechanical control, human-robot interaction, and rehabilitation, among others. In order to implement those models in MuJoCo, we developed a pipeline (see Appendix \ref{MyoSim:Pipeline}) to perform geometry transformations of bones and muscles attachment, moment arm optimization, and muscle force optimization.

A pipeline - MyoSim (see Appendix \ref{MyoSim:Pipeline}) - was developed to convert OpenSim musculoskeletal models into an equivalent MuJoCo model. This pipeline - based on three steps - enabled us to maximize the similarities between the newly developed MuJoCo models with the referencing OpenSim models, despite the difference in the wrapping surfaces, and the muscle dynamic definitions. It is worth noting that, these three steps are not specifically for the models presented in this benchmark, but are rather generalizable, i.e. then can be translated to other models. In addition, the proposed pipeline can operate based on data derived from cadaver data or from functional tasks, thereby allowing flexibility in building accurate musculoskeletal models directly in MuJoCo. These three steps are:

\begin{enumerate}
\item \textbf{Geometry Transfer:} The transformation of segment (bone) geometries, muscle attachment points, and wrapping surfaces, from an OpenSim model to a MuJoCo model. This step was based on extensions of previous work from \cite{ikkala2020}.
\item \textbf{Moment Arm Optimization:} The optimization of the 3D positions of side sites for the wrapping surfaces in MuJoCo, so that matching moment arms can be achieved with respect to the reference OpenSim model.
\item \textbf{Muscle Force Optimization:} The optimization of the muscle parameters in MuJoCo, so that matching muscle force generating-capacity can be achieved with respect to the reference OpenSim model.
\end{enumerate}

After rigorous modeling and validation, we built three models of varying complexities.

\begin{figure}[h!]
\centering
\includegraphics[width=\textwidth]{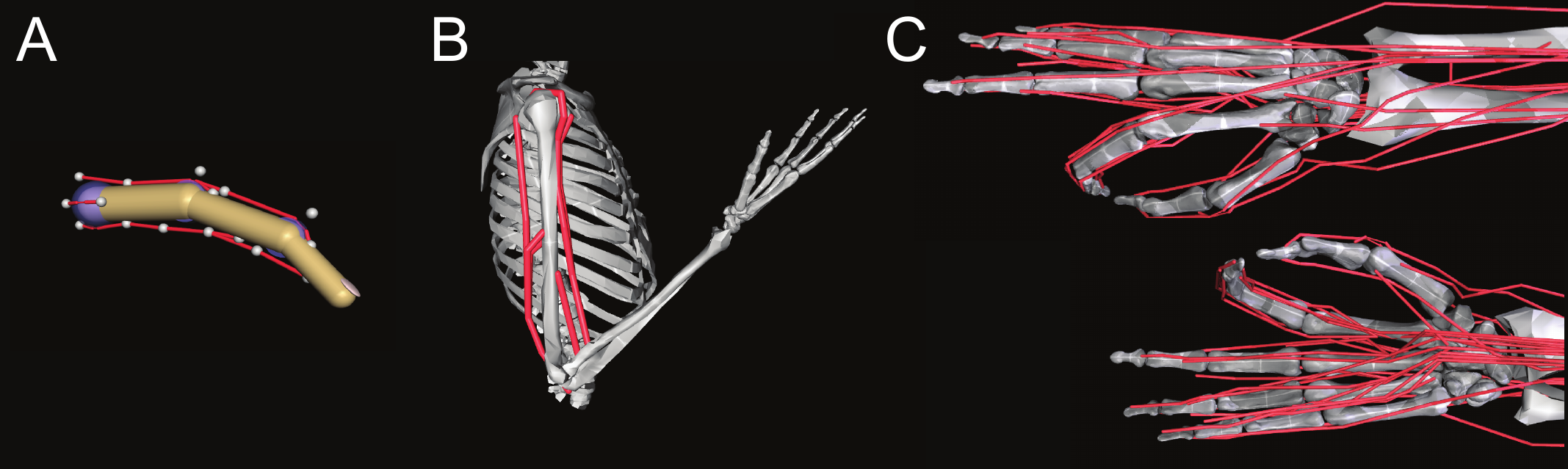}
  \caption{Musculoskeletal models included in MyoSuite. A - MyoFinger (4 joints - 5 muscles), B - MyoElbow: (1 joint - 6 muscles), C - MyoHand: (23 joints - 39 muscles).}
 \label{Fig:MuscSkeletal_Models}
\end{figure}

\textbf{{MyoFinger}}:\label{Sec:Finger_model} First, we implemented a simplified and intuitive model (based on \cite{xu2012design}) of a 4 Degree of Freedom (DoF) finger (MyoFinger, Figure \ref{Fig:MuscSkeletal_Models}A), which is actuated through a series of 5 simplified antagonistic muscle-tendon units. We also provide its robotic counterpart with simple torque actuators to facilitate the comparative investigation.

\textbf{{MyoElbow}}:\label{Sec:Elbow_model} A model of 1 DoF human elbow joint (Figure \ref{Fig:MuscSkeletal_Models}B). MyoElbow model is based on OpenSim's default testing arm model  (\cite{OpenSim_Delt2007, OpenSim_Seth2018}) and actuated using multiple agonist/antagonist pairs (3 flexors and 3 extensors).

\textbf{MyoHand}:\label{Sec:Hand_model} The dexterous human hand requires coordination of multiple highly redundant muscles, which have complementary and antagonistic effects on various joints. This more complex musculoskeletal model is comprised of 29 bones, 23 joints, and 39 muscles-tendon units (see section \label{NMSDetails} and Appendix \ref{'Table:MusclesNames'} for a detailed description of the muscles %
). This forearm-wrist-hand model (MyoHand, Figure \ref{Fig:MuscSkeletal_Models}C) was based on 2 popular OpenSim models: MoBL - human upper extremity model (\cite{Saul_2015} \cite{Mcfarlandl_2019}) - and the 2nd-Hand - for hand and fingers models (\cite{Lea_Asakawa_2015}).
Muscle attachments and range of operation in the original model were based on cadaveric studies \cite{an1983tendon, brand1993clinical, kuechle1997shoulder, langenderfer2004musculoskeletal, lieber1990architecture, murray1995variation, otis1994changes}.
Additionally, (a) our model also includes an Opponens Pollicis muscle for the critical role that it has in manual dexterity \cite{Karakostis_2021}, and (b) leverages functional studies \cite{wu2008analysis} to recover the force-length characteristics of the Opponens Pollicis, Radial Interosseous, and the Lumbrical Palmar (or Ulnar Interosseous), while defaulting to MoBL and 2nd-Hand for others.

\subsection{Tasks}
\label{sec:tasks}

Leveraging these musculoskeletal models, we built a series of tasks (see Figure \ref{Fig:DexterousTasks}) of varying difficulty. The task difficulty is varied along two axes: task-complexity, and task-non-stationarity. Task-difficulty has two variations (\textit{complexity} - easy/hard, and \textit{Reset} - Fix/Random/None), and task-non-stationarity has 8 (or 4 if tendon-transfer and exoskeleton assistance are not possible) variations. In total, in its current form MyoSuite consists of 204 tasks (Table \ref{all_tasks}): 9 task-families x 2 difficulties-levels x 3 Resets  x 8 (or 4) combinations of non-stationarity variations.

The definition of the tasks follows the ones presented in the github repository
\href{https://github.com/facebookresearch/myosuite}{\color{blue}https://github.com/facebookresearch/myosuite}.

\textbf{{Finger Joint Pose}:}
In the finger joint pose task, the MyoFinger model (Section \ref{Sec:Finger_model}, Figure \ref{Fig:MuscSkeletal_Models}A) needed to be controlled to reach a given pose. In the easy version (\textit{myoFingerPoseFixed-v0}) the target joint pose was fixed, while in the hard condition (\textit{myoFingerPoseRandom-v0}) the target pose was produced by a random configuration of the joints.

\textbf{{Finger Tip Reach}:}
In the finger tip reach task, the MyoFinger (Section \ref{Sec:Finger_model}, Figure \ref{Fig:MuscSkeletal_Models}B) tip needed to reach a fixed position - easy (\textit{myoFingerReachFixed-v0}) - or a random position - hard (\textit{myoFingerReachRandom-v0}) - in the working space.

\textbf{{Elbow Flexion}:}
The MyoElbow model (Section \ref{Sec:Elbow_model}, Figure \ref{Fig:DexterousTasks}C) was simplified to have only elbow rotations. Although it is not a  physiologically accurate model it can be a very simple model for troubleshooting initial control schemes.  There are 2 versions of this task: reaching a random posture either with one muscle (\textit{myoElbowPose1D6MFixed-v0}) or with 6 muscles (\textit{myoElbowPose1D6MRandom-v0})

\textbf{{Key Turn}}:
A key turn task (Figure \ref{Fig:DexterousTasks}E) where MyoHand finger movements need to be coordinated to rotate a key. There are two versions of this task of different difficulty: easy, to reach a half rotation of the key (\textit{myoHandKeyTurnFixed-v0}), and difficult, to reach full rotation of the key from different random position and rotations (\textit{myoHandKeyTurnRandom-v0})

\textbf{{Hand Tips Pose}:}
In this task, the MyoHand model (Section \ref{Sec:Hand_model}, Figure \ref{Fig:DexterousTasks}G) needed to be controlled to have the joint reach either a fixed pose
- easy (\textit{myoHandPoseFixed-v0-v0}) - or a random pose - hard (\textit{myoHandPoseRandom-v0-v0}).

\textbf{{Hand Tips Reach}:}
The fingers of the MyoHand model (Section \ref{Sec:Hand_model}, Figure \ref{Fig:DexterousTasks}F) needed to be controlled to have the fingers tips reach either a fixed position
- easy (\textit{myoHandReachFixed-v0}) - or a random position in their workspace - hard (\textit{myoHandReachRandom-v0}).

\textbf{{Hand Object Hold}:}
The MyoHand model (Section \ref{Sec:Hand_model}, Figure \ref{Fig:DexterousTasks}H) was controlled to maintain an object balanced in the hand without letting it fall. The object could be placed in a fixed position in the palm
- easy (\textit{HandObjHoldFixed-v0}) - or a random position - hard (\textit{myoHandObjHoldRandom-v0}).

\textbf{{Pen Twirling}}:
The MyoHand needed to be controlled to rotate a pen in the hand to reach a given orientation (indicated by the green object in the scene) a given orientation without making if falling off (Figure \ref{Fig:DexterousTasks}I). The complexity of this task is due to the intermittent contacts between the object and multiple fingers while trying to stabilize the object. There are two versions of this task of different difficulty: easy, target in a fixed orientation of the target (\textit{myoHandPenTwirlFixed-v0}), and difficult, target in a random orientation (\textit{myoHandPenTwirlRandom-v0}).

\textbf{{Baoding balls}}:
Finally, a highly dexterous baoding ball task (Figure \ref{Fig:DexterousTasks}J) involving simultaneous relative rotation around each other of two free-floating spheres over the palm. There are two versions of this task of different difficulty: easy, swap the positions of the ball (\textit{myoHandBaodingFixed-v1}) - and difficult - achieve continuous rotations of the balls (\textit{myoHandBaodingRandom-v1}).

\begin{table}[h!]
\begin{center}

\begin{tabular}{l|c|c|c|c|c|c|c|}
\cline{2-8}
 & \multicolumn{2}{|c|}{\textbf{Complexity}}     & \multicolumn{5}{|c|}{\textbf{Non-Stationarity}}         \\ \cline{2-8}
  & Easy/Hard & Reset & None & Sarcopenia & Fatigue & Tendon-transf. & Exo. \\ \hline
\multicolumn{1}{|l|}{Finger Joint Pose}  & \checkmark/\checkmark & F/R/N & \checkmark             & \checkmark                   & \checkmark                &           &               \\ \hline
\multicolumn{1}{|l|}{Finger Tip Reach}   & \checkmark/\checkmark  & F/R/N& \checkmark             & \checkmark                   & \checkmark                &             &             \\ \hline
\multicolumn{1}{|l|}{Elbow Joint Pose}   & -/\checkmark  & F/R/N & \checkmark    & \checkmark                   & \checkmark                &            &      \checkmark        \\ \hline
\multicolumn{1}{|l|}{Hand Key Turn}      & \checkmark/\checkmark  & F/R/N & \checkmark             & \checkmark                   & \checkmark                & \checkmark             &           \\ \hline
\multicolumn{1}{|l|}{Hand Joints Pose}   & \checkmark/\checkmark  & F/R/N & \checkmark             & \checkmark                   & \checkmark                & \checkmark             &           \\ \hline
\multicolumn{1}{|l|}{Hand Tips Reach}    & \checkmark/\checkmark  & F/R/N & \checkmark             & \checkmark                   & \checkmark                & \checkmark               &         \\ \hline
\multicolumn{1}{|l|}{Hand Object Hold}   & \checkmark/\checkmark  & F/R/N & \checkmark             & \checkmark                   & \checkmark                & \checkmark            &            \\ \hline
\multicolumn{1}{|l|}{Hand Pen Twirl}     & \checkmark/\checkmark  & F/R/N & \checkmark             & \checkmark                   & \checkmark                & \checkmark             &           \\ \hline
\multicolumn{1}{|l|}{Hand Baoding Balls} & \checkmark/\checkmark  & F/R/N & \checkmark             & \checkmark                   & \checkmark                & \checkmark             &           \\ \hline \\
\end{tabular}

\caption{List of different tasks in the MyoSuite with both the complexity (easy/hard and Reset - \textbf{F}ix, \textbf{R}andom, \textbf{N}one) and the non-stationarities (None, Sarcopenia, Fatigue, Tendon-transfer and exoskeleton) for each of them.}
\label{all_tasks}
 \end{center}
\end{table}

Next we detail various non-stationarities that are supported in \methodname~ tasks.

\subsection{Realistic non-stationary task-variations}
Muscle properties are constantly changing. These changes can be instantaneous - like for musculoskeletal injury or surgery - or can vary over a short time frame - like muscle fatigue or exoskeleton assistance. To study neuromuscular adaptation to non-stationarities due to these changes during the real work-life scenario, four different variations in muscle properties have been included: Sacropenia, tendon transfer, Fatigue, and exoskeleton assistance.

\textbf{Sarcopenia: }\label{sec:sarcopenia} Sarcopenia is a muscle disorder that occurs commonly in the elderly population (\cite{Cruz-Jentoft2019Sarcopenia}) and characterized by a reduction in muscle mass or volume. The peak in grip strength can be reduced up to 50\% from age 20 to 40 (\cite{Dodds2016GlobalData}). We modeled sarcopenia for each muscle as a reduction of 50\% of its maximal isometric force.

\textbf{Effect of Fatigue: }\label{sec:fatigue}Muscle Fatigue is a short-term (second to minutes) effect that happens after sustained or repetitive voluntary movement and it has been linked to traumas e.g. cumulative trauma disorder (\cite{chaffin2006occupational}). A dynamic muscle fatigue model (\cite{Ma2009AValidation}) was integrated into the modeling framework. This model was based on the idea that different types of muscle fibers have different contributions and resistance to fatigue (\cite{vollestad1997measurement}).  The implemented model mainly considers the current maximal force production of the muscle and is updated using the following equation:

\begin{equation}
F^{Max}_{upd}(t) = F^{Max} \exp\left({k_{fatigue}*\int^{t}_{0}-\frac{F^{Act}_{m}(u)}{F^{Max}}du}\right)
\end{equation}

With $F^{Max}_{upd}(t)$ the current updated maximal muscle force, $F^{Max}$ the maximal without fatigue, $F^{Act}_{m}$ the active part of the muscle force and $k_{fatigue}$ a fatigue coefficient with a value of 1.

\begin{figure}
    \centering
    \includegraphics[width=.47\textwidth]{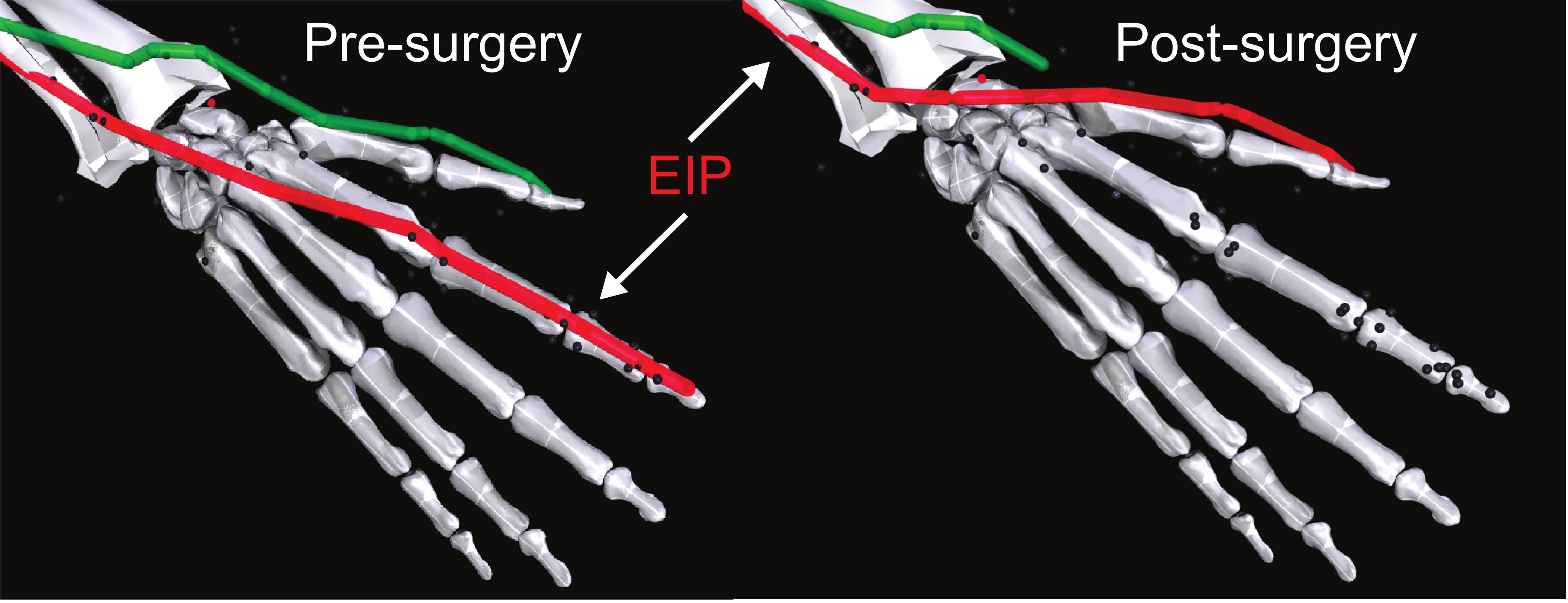}
      \caption{Tendon transfer. Left, EIP (in red - control the Index extensor) and EPL (in green - control the thumb extensor) muscles on an intact hand are shown. Right, the muscle path of EIP is re-routed on the EPL to control the activity of the thumb.}
     \label{Fig:TendonTransfer}
\end{figure}

\textbf{Tendon transfer pre/post surgery: }Contrary to muscle fatigue or sarcopenia that occurs to all muscles, tendon transfer surgery can target a single muscle-tendon unit. Tendon transfer surgery allows redirecting the application point of muscle forces from one joint DoF to another (see Figure \ref{Fig:TendonTransfer}). It can be used to regain functional control of a joint or limb motion after injury. One of the current procedures in the hand is the tendon transfer of Extensor Indicis Proprius (EIP) to replace the Extensor Pollicis Longus (EPL) (\cite{Gelb1995TENDONLONGUS}). Rupture of the EPL can happen after a broken wrist and create a loss of control of the Thumb extension. We introduce a physical tendon transfer where the EIP application point of the tendon was moved from the index to the thumb and the EPL was removed (see Figure \ref{Fig:TendonTransfer}).

\textbf{Exoskeleton assistance: }
\label{sec:HMI}
Human-Machine interaction simulations were generated as a use-case to illustrate how musculoskeletal model response to the exoskeleton assistance can be studied.
An elbow soft exoskeleton (Cable driven for example) was modeled as an ideal torque actuator perfectly aligned with the elbow joint with a weight of 0.101 Kg for the upper arm and 0.111 Kg on the forearm (Fig.  \ref{Fig:DexterousTasks}-D). The assistance given by the exoskeleton was a percentage of the biological joint torque, which represents a ideal version of the controller presented in \cite{durandau2019voluntary, durandau2017robust}.

\section{Experiments}
In this section, we first present results of the conversion pipeline to convert models from OpenSim to MuJoCo. Then, we present some baseline results on training reinforcement learning agents to solve some of the available tasks. Finally, we show some results when we train agents to solve tasks in presence of intrinsic non-stationary perturbation and while co-learning via human-robot interactions.

\subsection{Models validation}
\label{model_validation}
\begin{figure*}
\centering
\includegraphics[width=\textwidth]{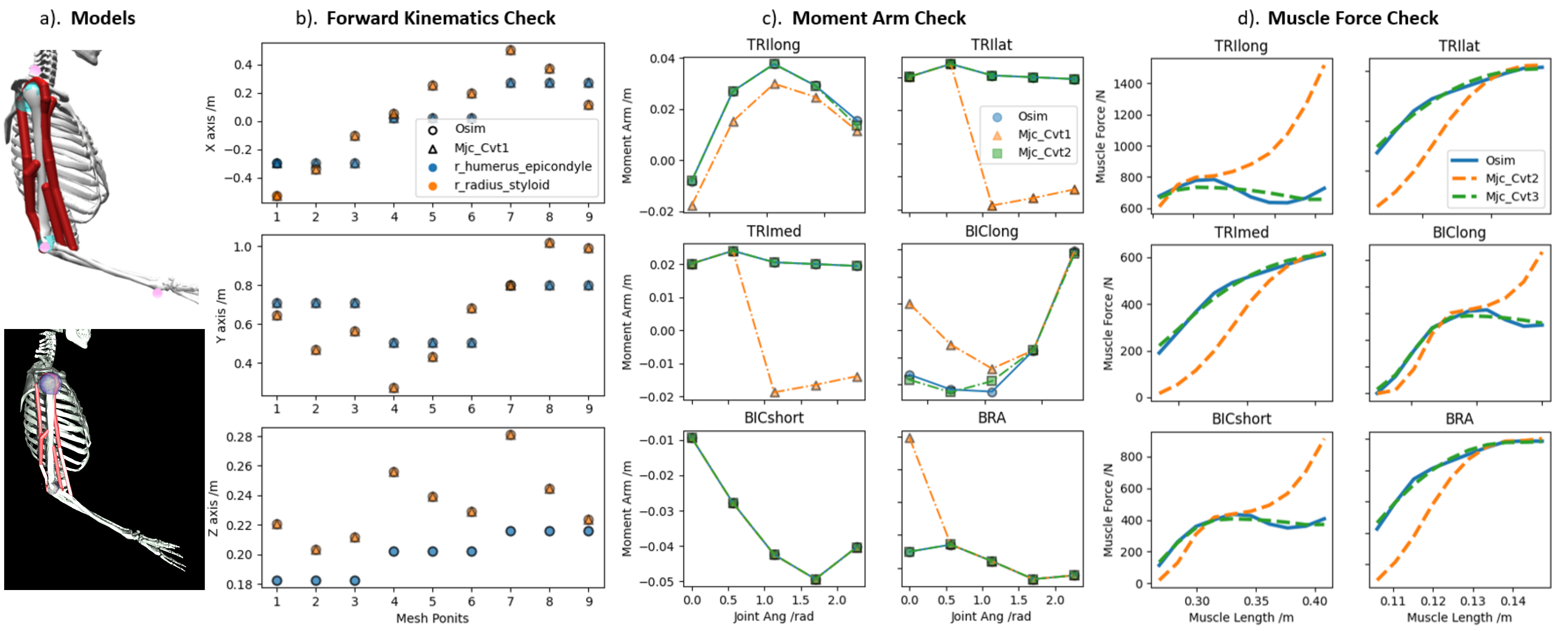}
  \caption{Validation between the converted MuJoCo and reference OpenSim models. a) The two models, top: OpenSim; bottom: MuJoCo. b) The forward kinematics validation. c) The moment arm validation. d). The muscle force validation. Muscle acronyms TRIlong - Triceps Long, TRIlat - Triceps Lateral,  TRImed - TricepsMedial, BIClong - Biceps Long, BICshort - Biceps Short , BRA - Brachialis}
  \label{Fig:validation}
\end{figure*}

A 2 joints 6 muscles (TRILong - Triceps Long, TRILat - Triceps Lateral,  TRImed - TricepsMedial, BIClong - Biceps Long, BICshort - Biceps Short , BRA - Brachialis) OpenSim elbow model \cite{OpenSim_Delt2007} is used to validate the conversion pipeline. The three steps validation is shown in Figure \ref{Fig:validation}. The first column shows the model appearances in OpenSim (top) and in MuJoCo (bottom). Second column indicates the matches of markers in forward kinematics checks. As shown in the plot, two selected markers have identical locations in both models, which means the match of geometry and joint definitions. The third column shows the results of moment arm validation. Compared to the OpenSim model, the converted MuJoCo model after Cvt1 ($Mjc\_Cvt1$) has large differences (the orange dash line with triangle markers), however after the second conversion step Cvt2, the moment arm differences between the OpenSim model and the MuJoCo model ($Mjc\_Cvt2$) is greately reduced (the green dashed lines are overlapping on the blue lines). The fourth column shows the validation of muscle forces. A large reduction of differences of muscle force was achieved also by applying the third step of the pipeline. Results of this last step are indicated by the matches between the blue solid line (OpenSim model), and the green dash line (converted MuJoCo model after the third step $Mjc\_Cvt3$). The yellow dash line (converted MuJoCo model shows the intermediate step $Mjc\_Cvt2$). Similar validation metrics and trends were observed in the full hand model as well.

\begin{figure}
\centering
\includegraphics[width=0.4\textwidth]{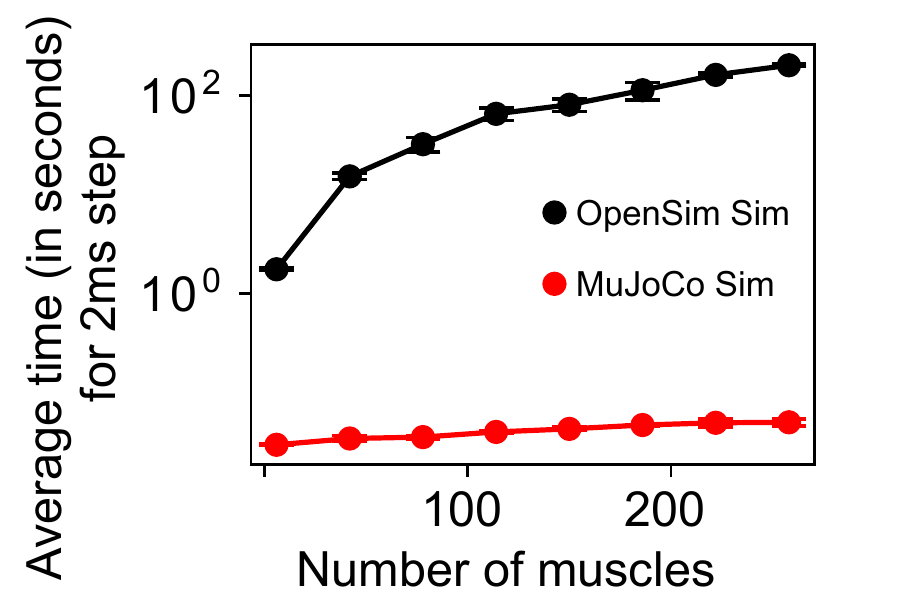}
  \caption{Performance comparison between MuJoCo and OpenSim with respect to the complexity of the model being simulated. }
  \label{Fig:speed_model_comparison}
\end{figure}

We compared the MuJoCo models against the correspondent OpenSim implementation. Muscle moment arm root mean square (RMS) differences between the MuJoCo model with respect to the Opensim model were $0.044 \pm 0.09 \% $ for the elbow model (see Figure \ref{Fig:validation}) and $0.38 \pm 0.57 \% $ for the hand model. Also, the RMS error in forces was $2.2 \pm 1.4 \%$  $F_{max}$ (OpenSim peak force) for the elbow model and $4.1 \pm 2.0 \%$ $F_{max}$ for the hand model (see Figure \ref{Fig:validation}). Those errors indicate that the MuJoCo models are anatomically and dynamically similar to the SOTA OpenSim model.
Forward simulations showed that MuJoCo models can be several orders of magnitude faster than OpenSim (see Figure \ref{Fig:speed_model_comparison}, from 60x to 4000x faster). By simulating an elbow model (6 muscles) where we iteratively replicated all muscles, it was possible to observe that the OpenSim computing time increased exponentially while the MuJoCo did not (see Figure \ref{Fig:speed_model_comparison}). This increase in efficiency is mostly the result of a simplified implementation of the muscle actuator in MuJoCo, which allows faster and more stable simulations.

\subsection{Baselines}
\begin{figure}[h!]
\centering
\includegraphics[width=0.75\textwidth]{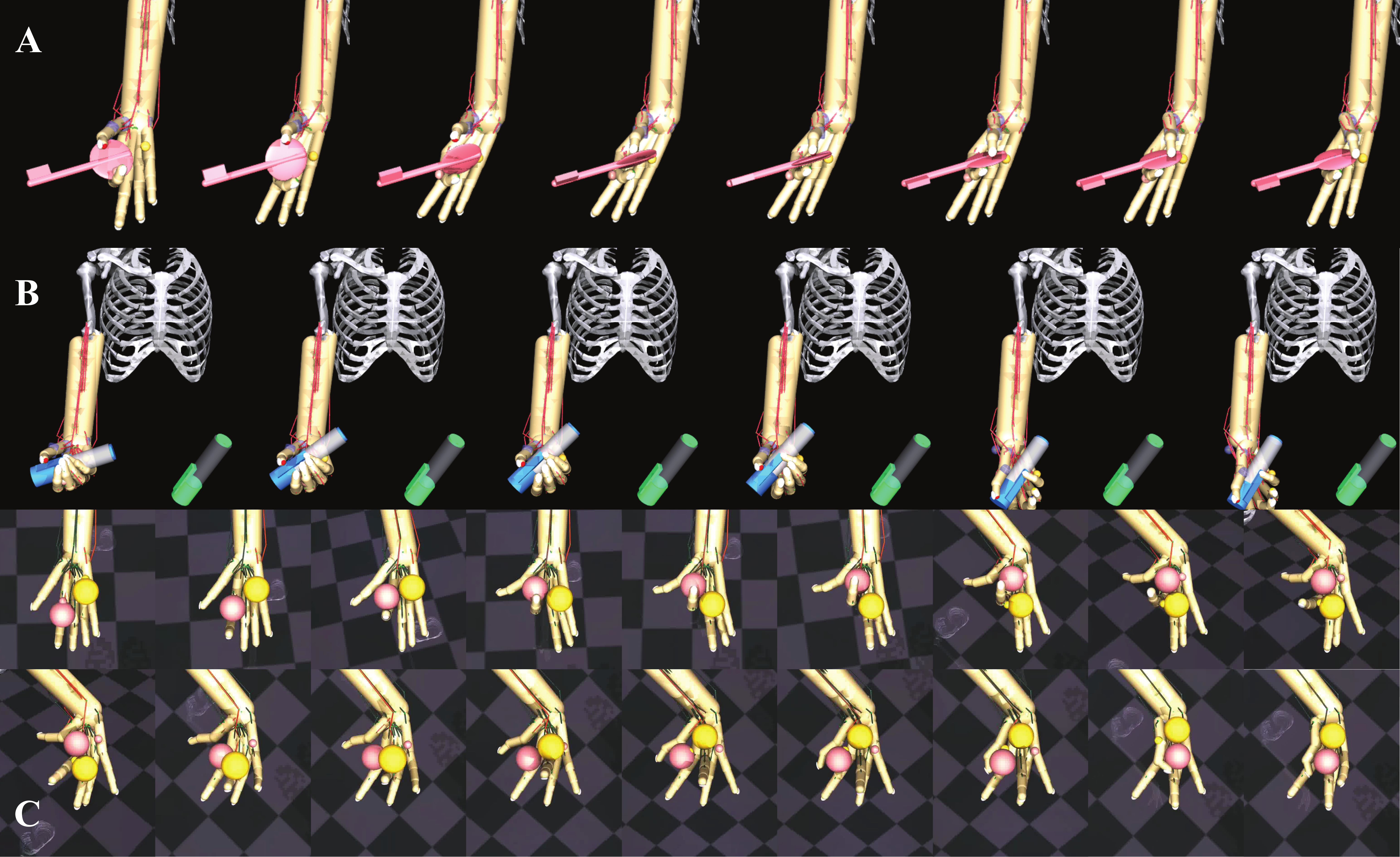}
  \caption{Frames from a selection of tasks. Complex wrist and finger coordination to solve the Key task (A) the Pen Twirl task (B) and the Baoding Balls task (C).}
  \label{Fig:Frames_Tasks}
\end{figure}

We present some baseline results obtained using Natural Policy Gradient (NPG, \cite{kakade2001natural} for a subset of the conditions available i.e. stationary easy and hard with Fix reset complexity. We chose this algorithm as it has been showing recently  SOTA results in dexterous manipulation tasks (\cite{Rajeswaran-RSS-18}).
In Figure \ref{Fig:RS_SR_tasks}, we show success rates for the different tasks up to 5M samples. More complex tasks e.g. Baoding ball can be solved, albeit with much higher sample complexity.  This success rate was reflected also in behavioral relevant movements. For example, in Figure \ref{Fig:Frames_Tasks}A, it is shown a sequence of snapshots of the solution of the key turning task. It is possible to see how the index and thumb activation is functional to the effective rotation of the key.
Also, in Figure \ref{Fig:Frames_Tasks}B a sequence of snapshots of the Pen Twirl task is shown. Indeed, in this task, it is visible how the coordination of wrist and hand joints make it possible to rotate the pen in the hand to reach the desired position. Finally, in Figure \ref{Fig:Frames_Tasks}C a sequence of snapshots of the Baoding tasks show the complex coordination between wrist and fingers to move and rotate the balls against each other. It has to be noticed that solving the Baoding ball tasks was achieved with a greater number of samples i.e. more than 70M, and only occasionally i.e. one seed out of 5.

\begin{figure}[h!]
\centering
      \centering
        \includegraphics[width=.75\textwidth]{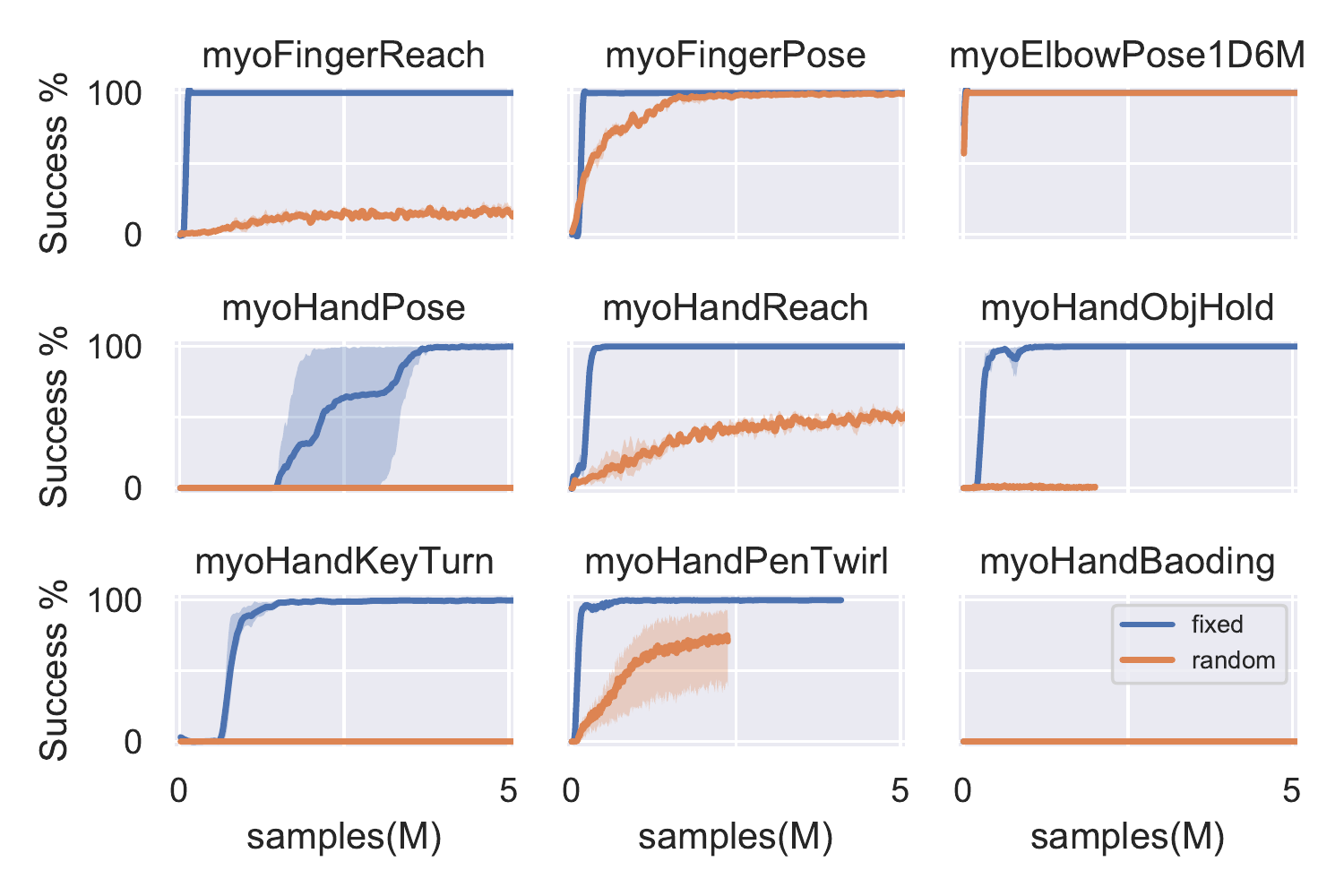}
        \caption{Tasks performance in the easy (fixed) and the hard (random) conditions with fixed reset.}\label{Fig:RS_SR_tasks}
       \label{Fig:RS_SR_tasks}
\end{figure}

\subsection{Intrinsic non-stationary perturbations}

While extrinsic perturbation can be easily added to each task, we show results testing the above policies with a series of intrinsic non-stationary perturbations compatible with real-life scenarios.

\subsubsection{Control in presence of intrinsic non-stationary perturbation: synergistic muscles contributions}
\begin{figure}[h!]
    \centering
    \includegraphics[width=\textwidth]{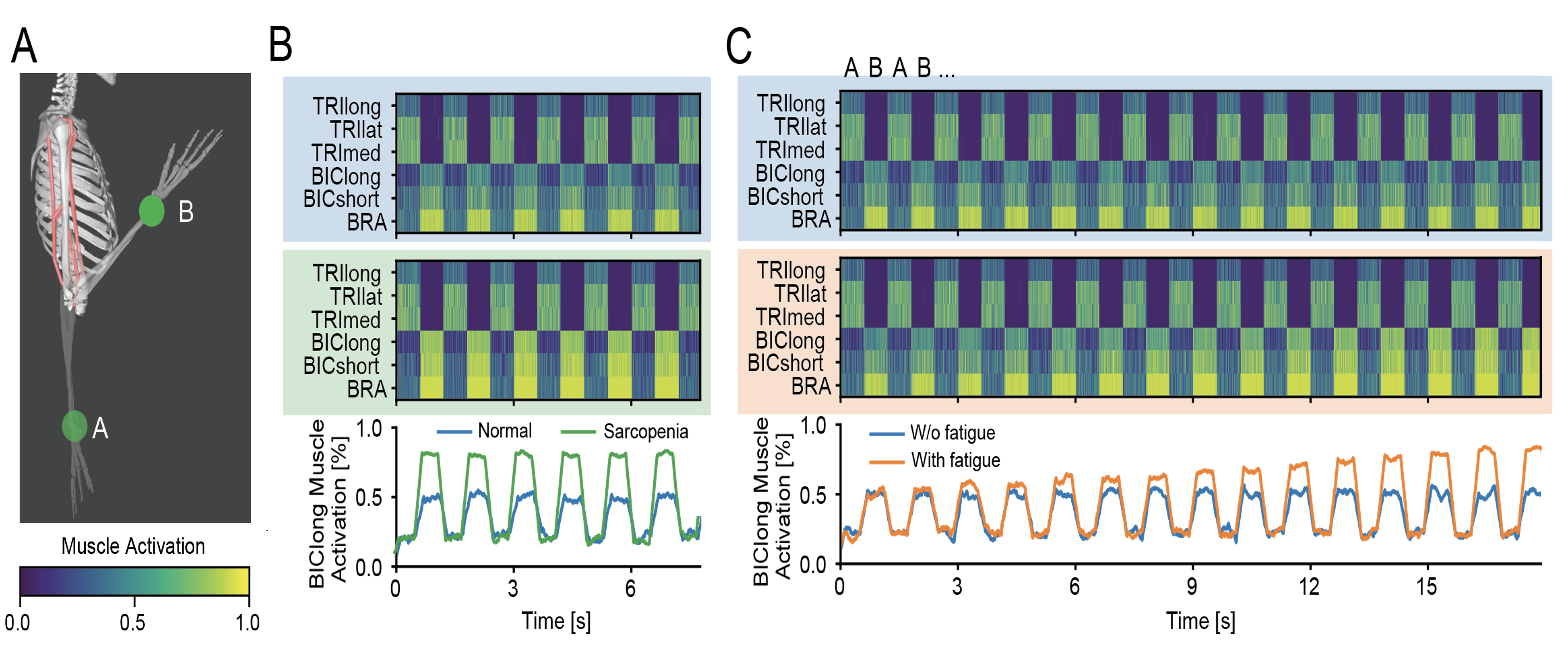}
      \caption{Control Policy with muscle sarcopenia (weakness) and fatigue. (A) Experiment design: Train a policy to reach random position of the wrist and test with a task of making alternate reaching movements between two elbow angle. (B) Test of the effects of Sarcopenia. Top-row: muscle activations for 6 arm muscles (3 agonists and 3 antagonists). Middle row: muscle activations in presence of sarcopenia. All synergistic flexor muscles (BRA - brachioradialis, BICLong - Biceps Long, BICShort - Biceps short) increase their activation to compensate for the reduced force. Bottom row: a trace of the  activation only for the Biceps Long muscle. (C) Fatigue: Without fatigue weaker activation of the Biceps Long and Biceps Short where enough to reach the B-posture, in presence of fatigue,  stronger activations from the three synergistic muscles are needed.}
    \label{Fig:Weakness_and_Fatigue6Muscles}
\end{figure}

We tested a policy trained on a 1D elbow flexion task to reach random targets in its operational space on alternated movements between points A and B (see Figure \ref{Fig:Weakness_and_Fatigue6Muscles}A).

First, we study how sarcopenia (muscle weakness) affects the control of movement.  In presence of Sarcopenia (Figure \ref{Fig:Weakness_and_Fatigue6Muscles}B) the Brachioradialis - which in normal conditions can solve the task without the support of other muscles - needs stronger activations and the support of synergistic muscles (BICLong - biceps longus - and BICShort - biceps short) to solve the task.

Second, we investigated the effect of muscle fatigue (introduced in section \ref{sec:fatigue}). In this case, the loss of muscle power is progressive over time. In the same alternated movements elbow tasks, we observe a gradual contribution of synergistic muscles to compensate for the muscle force loss (Figure \ref{Fig:Weakness_and_Fatigue6Muscles}C).

\subsubsection{Accidents (Tendon tear)}
\begin{figure}[h!]
    \centering
    \includegraphics[width=\textwidth]{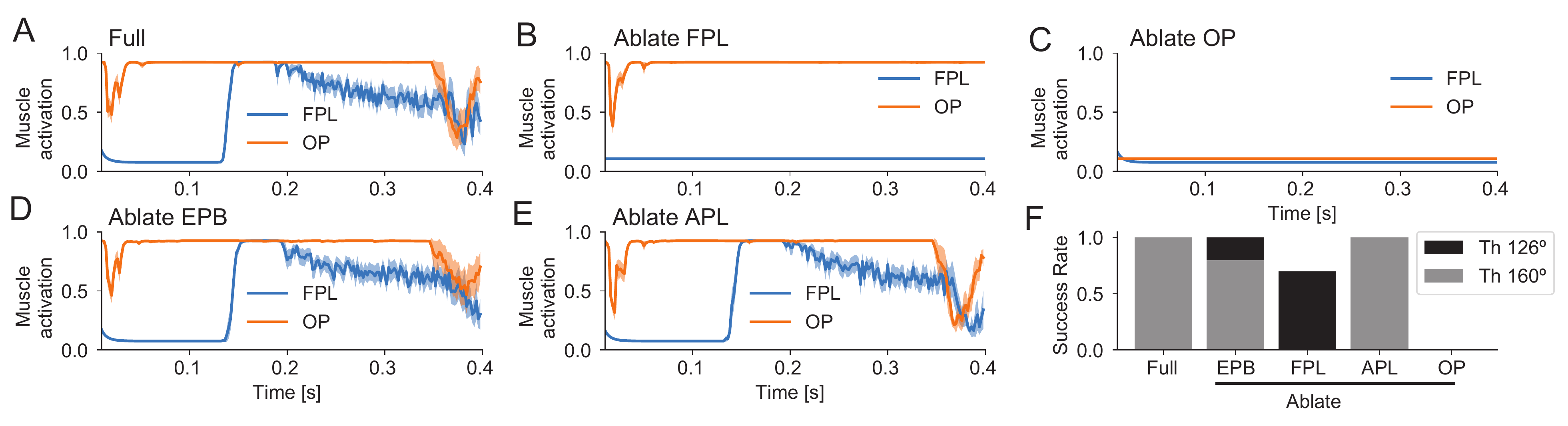}
      \caption{Effect of Tendon tear on the key-turning task. In this task a policy was trained to solve the key turning task. After training, selectively a different thumb muscle was removed for each experiment: 2 flexors (Flexor Pollicis Longus (FPL), Opponens Pollicis (OP)) and 1 extensor (Extensor Pollicis Brevis (EPB)) and 1 abductor (Abductor Pollicis Brevis (APL)), and the task was repeated 10 times. Normal thumb (Panel A). When FPL was ablated (Panel B), OP muscle continued its activation to accomplish the task while during ablation of OP (Panel C), FPL muscle did not show any activation. Ablation of Extensors muscles did not have an effect on the other muscles activation patterns (D-E). Finally, summarized in Panel F are the success rates in turning the key to reach a threshold of 126$^{\circ}$ or 160$^{\circ}$ rotation.}
      \label{Fig:Ablation_KeyTurn}
\end{figure}

Next, in order to study the effects of tendon tear, a policy was first trained to solve the Key-Turn task and then challenged with the selective damage(tear) of different thumb muscles (Figure \ref{Fig:Ablation_KeyTurn}). %
While there are redundancies (not every tear crippled ability), we found that  FPL and OP are critical for the correct solution of the key task. In absence of OP the FPL is able to compensate. But when OP is torn FPL cannot compensate resulting in reduced key rotations.

\subsubsection{Control in presence of tendon transfer}
\begin{figure}[h!]
    \centering
    \includegraphics[width=\textwidth]{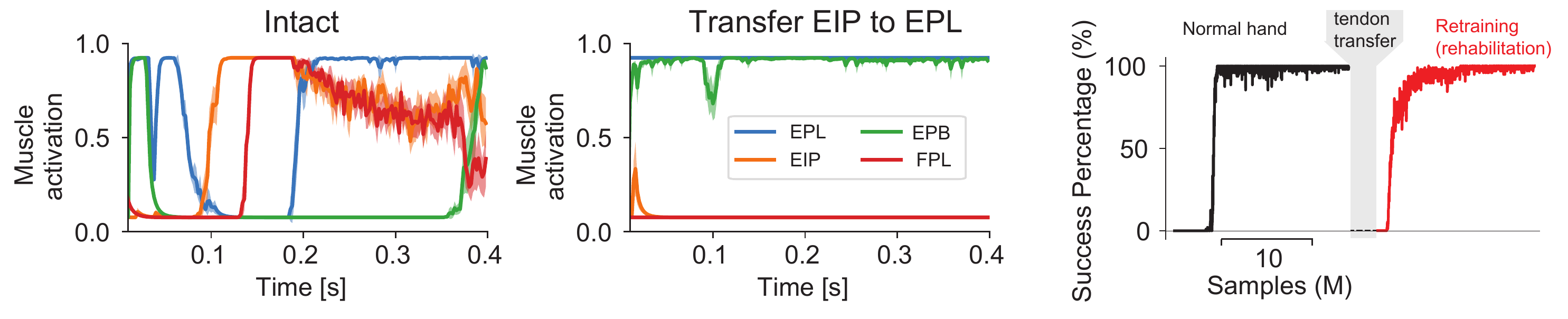}
      \caption{Index to thumb tendon transfer. In this experiment we first test a key-turning task on an intact hand and then we operated a tendon transfer: activity of the muscle Extensor Indicis Proprius (EIP) was redirected to activate muscle Extensor Pollicis Longus (EPL). In the left panel, thumb muscles (EPL, EIP, Extensor Pollicis Brevus (EPB), and (Flexor Pollicis Longus (FPL)) are shown (average over 10 repetitions) as the are orderly recruited to perform the task. After EIP to EPL remapping - center - the normal ordered recruitment is lost and the task cannot be any longer performed. The policy needs to be retrained in order to solve the task (right).}
    \label{Fig:TendonTransfer_KeyTurn}
\end{figure}

Finally, a tendon transfer - EIP muscle was routed to replace the EPL muscle - was performed to test the ability of the policies to compensate for action re-mapping due to tendon surgery. After the surgery, the major thumb muscles needed to compensate for the different activation space (Figure \ref{Fig:TendonTransfer_KeyTurn} and, %
a previously trained policy was unable to solve the task. Indeed it was necessary an extensive additional training to control the thumb after tendon transfer. This is typically observed in patient undergoing extensive physiotherapeurical sessions to re-learn the control of the thumb (\cite{wangdell2016early}).

\begin{figure*}[h]
\centering
\includegraphics[width=1\textwidth]{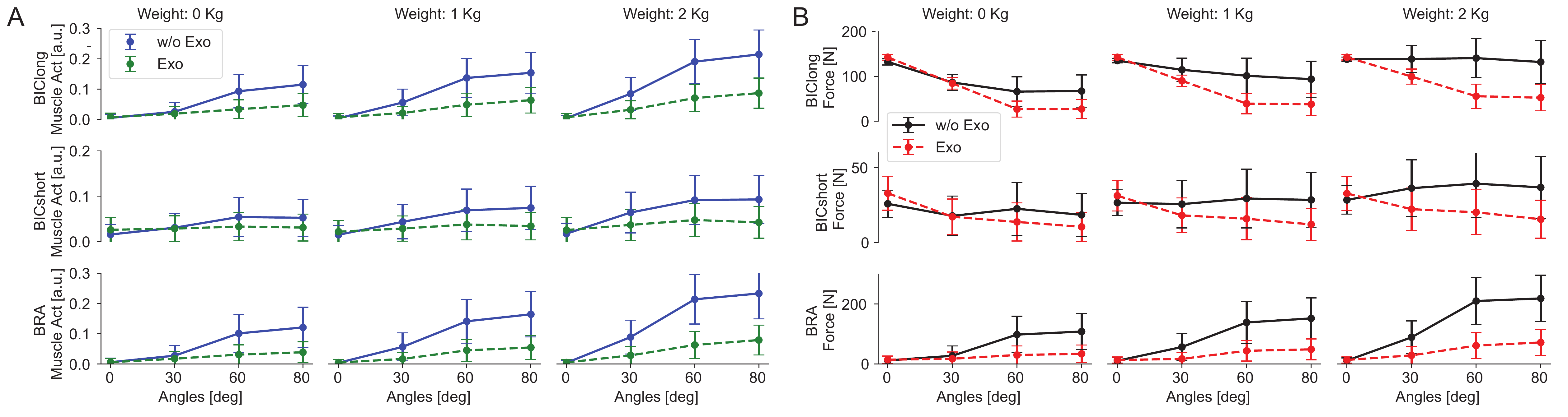}
 \caption{Effects of loads on Exoskeleton assisted reaching. A policy was trained in a reaching task of exoskeleton assisted elbow flexion to reach targets at $0\deg$, $30\deg$, $60\deg$, $80\deg$ and hold the position for 2 seconds. A - Flexor muscles activations without (blue) and with (green) exoskeleton assistance at different target positions. With exoskeleton assistance the muscle activation required to reach the target angles was less. B - Same analysis as for A but on forces. }
\label{Fig:Exo_Weigths}
\end{figure*}

\subsection{Human Robot interaction}
In the MyoElbow model, we locked shoulder movements and we added an exoskeleton (see Appendix \ref{sec:HMI}). Simulation of human robot interaction were realized for two experiments: I) flexion extension (between 0 - 30, 0 - 60 and 0 - 80 degrees with 2 sec hold time) with an healthy model with different weights on the hand (task inspired by \cite{lotti2020adaptive}) and II) static with the elbow at 90 degrees with an healthy, Sarcopenia (see section \ref{sec:sarcopenia}) and fatigue (see section \ref{sec:fatigue}) conditions. The trials were repeated for the standard arm model with and without the exoskeleton support presented in section \ref{sec:HMI}.

We trained a joint policy of exo and muscles by means of a Natural Policy Gradient (NPG) \cite{kakade2001natural} algorithm in the natural (healthy) condition without any weight on the hand to reach random targets in the range of elbow flexions between 0 and 130 degrees of the elbow.

Figure \ref{Fig:Exo_Weigths} shows the effects of the reaching task (Experiment I)  without and with exoskeleton assistance when a load is applied on the hand. As expected the reaching is minimally impacted when the exoskeleton is not functional but there is no additional perturbation. When a weight is applied the exoskeleton is able to recover the original position with about 5 degrees error (distance of joint angles, mean $\pm$  std, 0 Kg: 4.95 $\pm$ 3.42, 1 Kg: 4.54 $\pm$ 3.12, 2 Kg: 4.34 $\pm$ 3.43). This is obtained both with a ~60\% reduction of both the muscle activation (Figure \ref{Fig:Exo_Weigths}A, BIClong 60\%, BICshort 54\%, BRA 66\%) and forces (Figure \ref{Fig:Exo_Weigths}B, BIClong 60\%, BICshort 58\%, BRA 67\%) at the weigth of 2 Kg.

\begin{figure}[h!]
    \centering
    \begin{minipage}{.49\textwidth}
    \centering
    \includegraphics[width=.8\textwidth]{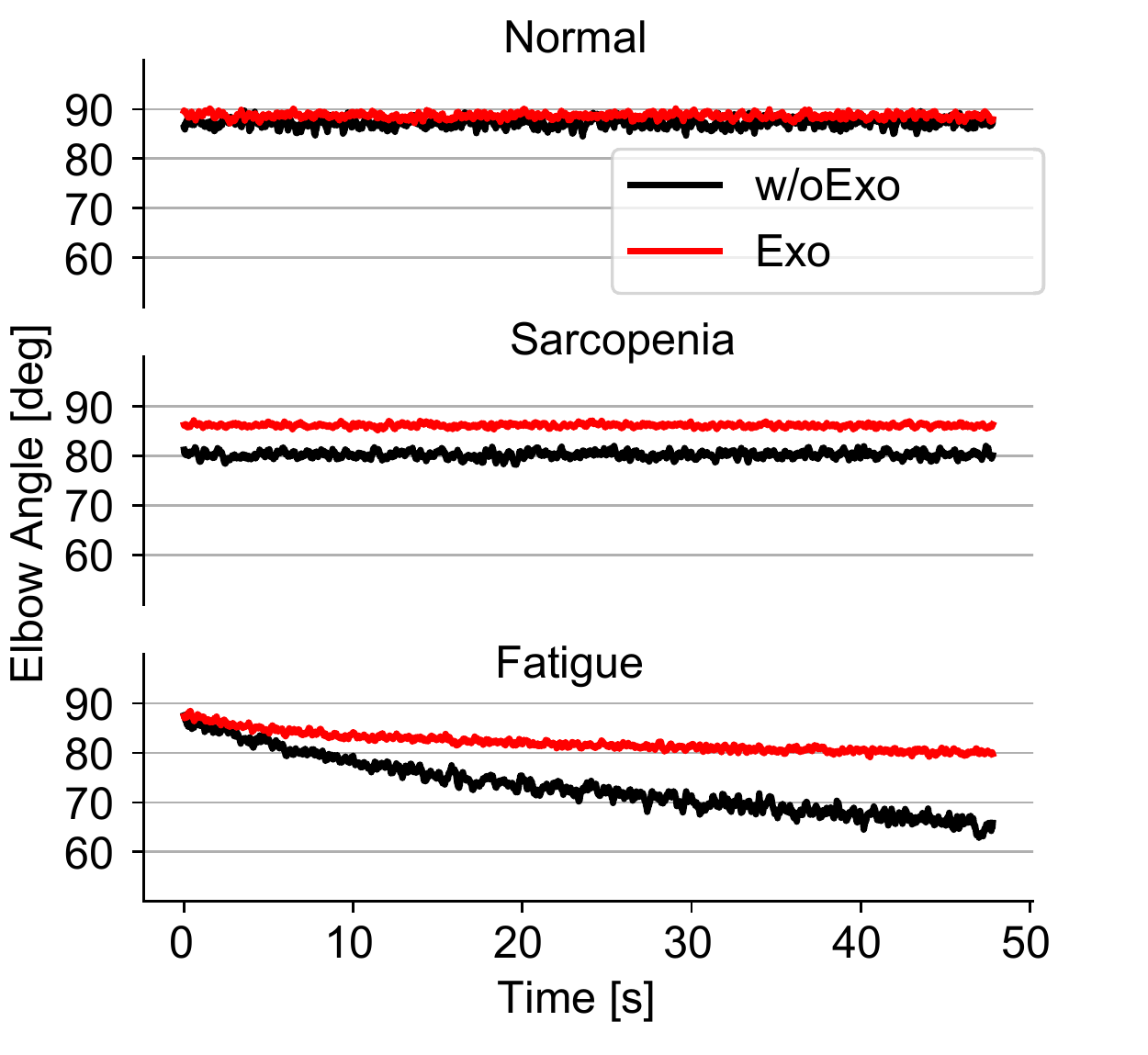}
     \caption{Effects of Sarcopenia and Fatigue on Exoskeleton assisted reaching. The same policy trained for exoskeleton assisted reaching was tested on holding a $90\deg$ angle elbow posture. The three rows show the effect of Normal, Sarcopenia and Fatigue respectively on holding the position. During Sarcopenia the muscle loss of force (weakness) is almost completely recovered with the exoskeleton. Likewise for fatigue, the exoskeleton compensate partially for the loss of force. Nevertheless, since the policy was not trained to compensate for those conditions, as results it is not able to drive the exoskeleton to a complete recovery of the function.}
    \label{Fig:Exo_Sar_Fat_time}
     \end{minipage}
    \hfil
    \begin{minipage}{.49\textwidth}
    \centering
    \includegraphics[width=\textwidth]{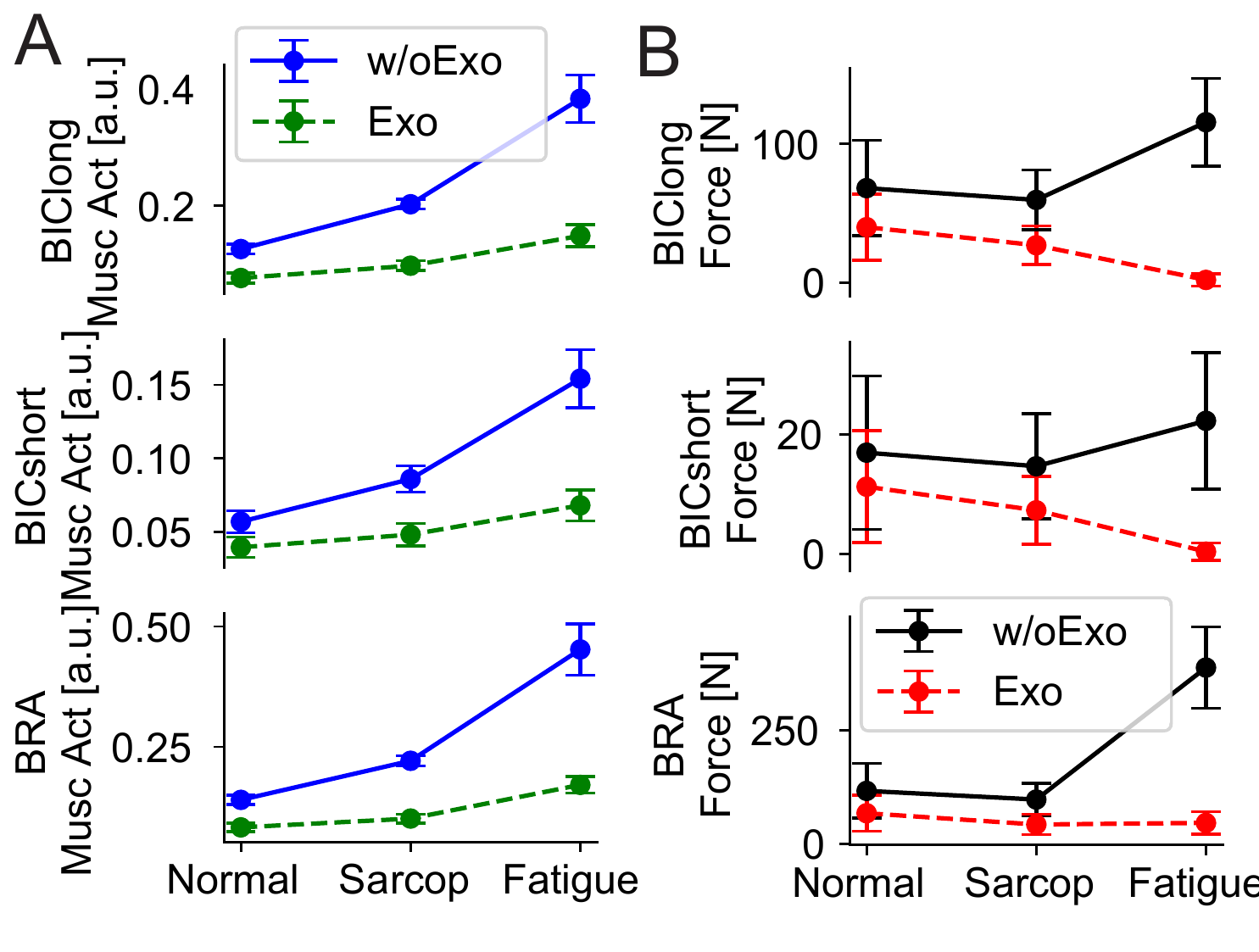}
     \caption{Effects of Sarcopenia and Fatigue on Exoskeleton assisted holding. Same task and policy as for \ref{Fig:Exo_Sar_Fat_time}. A - Flexor muscles activations without (blue) and with (green) exoskeleton assistance with the different conditions. With exoskeleton assistance the muscle activation required to compensate for the functional deficit is decreased. B - Same analysis as for A but on the forces. }
    \label{Fig:Exo_Sar_Fat}
     \end{minipage}
\end{figure}

Figure \ref{Fig:Exo_Sar_Fat_time} and Figure \ref{Fig:Exo_Sar_Fat} show the effects of holding a static position without and with exoskeleton in the presence of intrinsic perturbation to the muscle models in the form of sarcopenia (muscle weakness) and fatigue. As expected the holding is minimally impacted when the exoskeleton is not functional (distance of joint angles, mean $\pm$ std: 4.12 $\pm$  2.97) but there is no additional intrinsic perturbation (first row of Figure \ref{Fig:Exo_Sar_Fat_time}). When intrinsic perturbations are present, the use of the exoskeleton is able to partially recover for those conditions (mean $\pm$ std, Sarcopenia 6.03 $\pm$  2.97, Fatigue 11.89 $\pm$ 4.03). This is obtained both with a reduction of the muscle activation (Figure \ref{Fig:Exo_Sar_Fat}A, reduction without vs with exoskeleton of Normal, Sarcopenia and Fatigue for BIClong: 40\%, 52\%, 61\%, BICshort: 30\%, 44\%, 56\%; BRA: 41\%, 54\%, 62\%) and forces (Figure \ref{Fig:Exo_Sar_Fat}B, reduction without vs with exoskeleton of Normal, Sarcopenia and Fatigue for BIClong: 41\%, 55\%, 98\%; BICshort 34\%, 50\%, 98\%, BRA: 43\%, 56\%, 88\%).

\section{Discussion and Conclusions}
Here, we have proposed a new framework of physiologically realistic and computationally efficient models and tasks to study human motor control. The proposed models include realistic non-stationarities to simulate real-life scenarios such as muscle fatigue, sarcopenia, and tendon transfer. The suite of tasks entails realistic highly skilled manipulations. We hope that the wide range of complex and realistic challenges that the proposed platform affords, provides benchmarks that catalyze robot control in the same way that challenging datasets and benchmarks did for computer vision and natural language. This benchmark will provide biologically relevant problems where task success and physiological representations might differ. Actually, because activation of the musculoskeletal models can be directly related to a (normalized) intramuscular activity recorded in human subjects, it will be possible to validate experimentally the learning solutions.

The basic policies trained already showed physiologically relevant behaviors. First, we observe an automatic adaptation of muscles to leverage antagonistic effects of flexors and extensors. This was clearly visible with the elbow model where those effects could be more easily isolated. Second, co-contractions evolve naturally to compensate for changes in muscle properties e.g. sarcopenia and fatigue. Nevertheless, those compensations are not enough to replace muscles like the Opponens Pollicis which has unique functions for hand manipulations.

Finally, it is worth mentioning that the implemented models are only the first iteration of an approximation of the musculoskeletal system that will need further development and validation. Indeed, both the tasks and physiological changes in muscle properties are a subset of the possible changes that can be considered.

All in all, we would like this work to facilitate cross-pollination and catalyzation of new ideas between different communities. The ML community will have more challenging tasks with 3rd order dynamics (which are uncommon in ML/RL benchmarks)\ and the proposed non-stationarity provides a more physiological realistic use-case to challenge their algorithms.  The biomechanics community will have the possibility to afford a new platform where contact-rich interactions at scale can be studied. Finally, the robotic community will have a biological system to develop strategies for. Overall, this will allow for the creation of realistic simulation opening the ways to \textit{in silico} trials of humans, robots, and their interaction.

\section*{Acknowledgments}

{H.W., G.D., and M.S. received funding by the European Research Council (ERC) under the European Union's Horizon 2020 research and innovation programme as part of the ERC Starting Grant INTERACT (Grant No. 803035) as well as part by the Horizon 2020 ICT-10 Project SOPHIA (871237).}

\newpage
\printbibliography %

\newpage
\appendix

\textbf{\huge Appendix}

\section{MyoSim: A pipeline to generate MuJoCo musculoskeletal models} \label{MyoSim:Pipeline}

In this section, the musculoskeletal model conversion tool is explained in detail.

\subsection{Musculoskeletal model conversion tool}
\label{sec:conversion}
\begin{figure}[h]
\centering
\includegraphics[width=.5\textwidth]{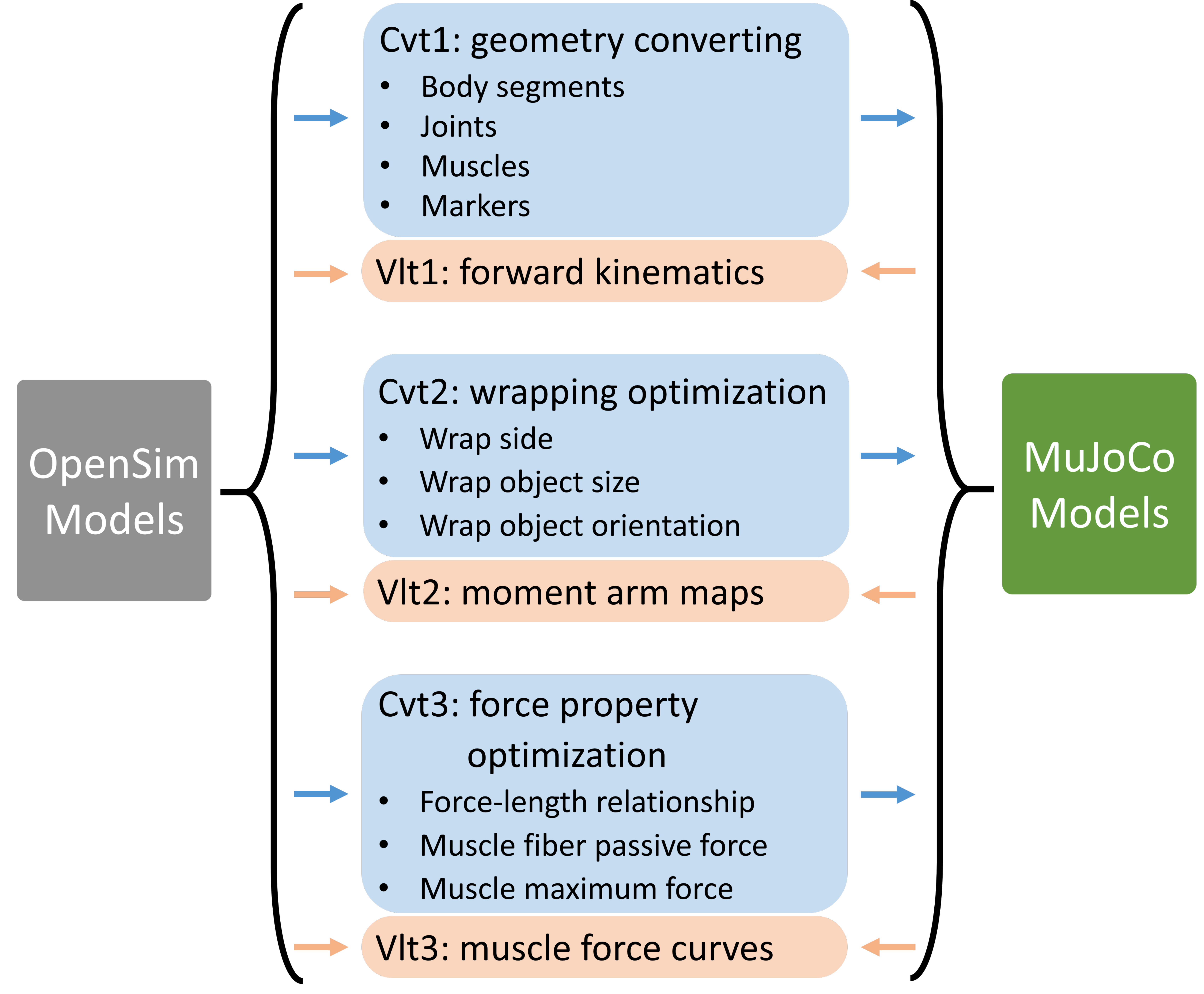}
  \caption{Three-steps conversion pipeline}
\label{fig:pipeline}
\end{figure}

We used physiologically validated and widely adopted OpenSim musculoskeletal models as reference. Then, a layer-by-layer automatic conversion pipeline was developed to convert those OpenSim  models into MuJoCo. The pipeline consisted of three major conversion (Cvt) steps, as shown in Figure \ref{fig:pipeline}. A validation (Vlt) module followed each conversion step to match the converted MuJoCo model and the reference OpenSim model. A short description of them is listed below, followed by the detail explanations in the subsections.

\begin{enumerate}
    \item \textbf{Cvt1: geometry conversion.} Parse kinematics details (body segment lengths, joints, muscles-tendon pathways, etc)\cite{ikkala2020} and fiducial markers from OpenSim model into MuJoCo model.

    \textbf{Vlt1: forward kinematics.} Check that salient positions along the kinematic chain are matched between the two models at different joint configurations.

    \item \textbf{Cvt2: wrapping optimization.} Optimize wrapping constraints to guarantee that the muscle pathways are physiologically reasonable.

    \textbf{Vlt2: moment arm maps.} Check that moment arms of each muscle at each joint angle are matched between models.

    \item \textbf{Cvt3: force property optimization.} Optimize muscle force parameters in MuJoCo to approximate OpenSim model's force curves.

    \textbf{Vlt3: muscle force curves.} Check force-length-activation differences between  the two models.
\end{enumerate}
\vspace{0.2cm}
\noindent \textbf{Geometry Conversion:} \label{geom_transfer}\\
OpenSim and MuJoCo geometry structures are simply defined based on their relative position in a similar hierarchical (XML) format. An open-access conversion code was created by Ikkala et al \cite{ikkala2020} that parsed the bodies, joints, and muscles into the MuJoCo modelling format. We extended their code by adding in the conversion 1) markers and 2) wrapping surfaces for muscles to avoid collisions with bones during movements. The forward kinematics module (\textit{Vlt1}) checks the differences of the selected marker locations at different joint configurations (see the Results section \ref{model_validation}).

\vspace{0.2cm}
\noindent \textbf{Wrapping Optimization:} \label{geom_transfer}\\
OpenSim and MuJoCo have different ways of defining how muscles wrap over wrapping surfaces. In OpenSim, three wrapping methods were defined: MidPoint, Axial, Hybrid \cite{delp2007opensim, seth2018opensim}. Nevertheless, in MuJoCo, muscles can wrap over or can be forced to pass through a surface. In order to guarantee similar forces produced by a muscle at each joint, an optimization was developed. We adapted both the location where a muscle wraps on the surface in MuJoCo i.e. \textit{`side site'} and the dimensions of the wrapping objects (surfaces such as ellipsoids and torus, which are missing in MuJoCo, were approximated with other surfaces).

The following optimization was developed:
\begin{equation}\label{MomentArmOpt}
\begin{aligned}
\noindent
& \textrm{\textbf{Find:} wrapping sides, sizes \& orientations:} \\
& \hspace{0.3 in} X = \{[L_1, ..., L_n], [S_1, ..., S_p], [R_1, ..., R_p]\} \\
& \textrm{\textbf{To minimize:} objective function:} \; \\
& \textrm{\hspace{0.3 in}} F = \sum_{m=1, j=1}^{M, J} (d^{osim}_{m,j}(Q) - d^{mjc}_{m, j}(Q, X))^2\\
& \textrm{\textbf{When:} iterate all joint angle meshes} \;\\
& \textrm{\textbf{Subject to:} model kinematics; parameter bounds}
\end{aligned}
\end{equation}

\noindent where, $L = (x, y, z)$ represents the location of side site; $S=(r, l)$ represents the size of wrapping e.g. cylinder dimensions;  $R=(\alpha, \beta, \gamma)$ are the Euler angles of wrapping surface orientation; $d$ represents the moment arms; $Q=[q_1,...q_J]$ represents joint angles; $M$ represents the total number of muscles; $J$ represents the total number of joints

Simulated annealing (SA) \cite{van1987simulated} was used as the optimization method to minimize the differences of moment arms i.e. errors, between the converted MuJoCo model and the referencing OpenSim model for all muscles at all joints angles. Details of this can be found in the validation section \ref{model_validation}.

\vspace{0.2cm}
\noindent \textbf{Force Property Optimization:}\label{force_optim}\\
The OpenSim and the MuJoCo platforms also use different methods for modelling muscle force properties. MuJoCo does not consider elastic tendons, fiber pennation angles, while OpenSim has both of them. OpenSim uses physical lengths for the definition of optimal fiber and tendon slack lengths, whereas, MuJoCo uses normalized value. Even though the MuJoCo muscle model is a simplified version of the OpenSim muscle model, it is still possible to generate similar force-length-velocity relationships. However, the muscle property parameters needs to be optimized, instead of directly mapping from the OpenSim model. The following optimization was developed to match the muscle force properties:

\begin{equation}\label{MuscleForceOpt}
\begin{aligned}
& \textrm{\textbf{Find:} muscle force parameters:} \\
& \textrm{\hspace{0.5 in}} X = \{l_{min}, l_{max}, fp_{max}, f_{max} \} \\
& \textrm{\textbf{To minimize:} objective function:} \; \\
& \textrm{\hspace{0.5 in}} F = \sum_{m=1}^{M} (f^{osim}_{m}(l_m, a) - f^{mjc}_{m}(l_m, a, X))^2\\
& \textrm{\textbf{When:} iterate all muscle length $l_m$ and activation $a$} \;\\
& \textrm{\textbf{Subject to:} muscle dynamic; parameter bounds}
\end{aligned}
\end{equation}
\noindent where, $l_{min}$ is the minimal normalized muscle length when active fiber force reaches zero; $l_{max}$ is the maximum normalized muscle length when active fiber force reaches zero; $fp_{max}$ is the normalized passive fiber force when muscle length equal to $l_{max}$; $f_{max}$ is the absolute value of maximum muscle force; $f(l_m, a)$ represents the muscle forces when the muscle has an activation of $a$ and at the length of $l_m$.

Differential evolution (DE) \cite{price2013differential} was used as optimization method. Muscle activation dynamics and the force-velocity relationship are identical between the OpenSim and MuJoCo models, as a result, they do not need to be optimized, but can be directly transferred. Validation of this step is the errors of force-length-activation maps between the OpenSim and MuJoCo models. More details of it can be found in the validation section \ref{model_validation}.

\begin{table}
\begin{center}
    \begin{tabular}{ c c c c c}
    \textbf{Label} & \textbf{Muscle} & \textbf{Number of muscles}  \\
     \hline
    ECRL & Extensor Carpis Radialis Longus & 1x \\
    ECRB & Extensor Carpis Radialis Brevis & 1x \\
    ECU & Extensor Carpi Ulnaris & 1x \\
    FCR & Flexor Carpi Radialis & 1x \\
    FCU & Flexor Carpi Ulnaris & 1x \\
    PL  & Palmaris longus & 1x \\
    PT  & Pronator teres & 1x \\
    PQ  & Pronator & 1x \\
    EIP & Extensor Indicis Proprius & 1x \\
    EPL & Extensor Pollicis Longus & 1x \\
    EPB & Extensor Pollicis Brevis & 1x \\
    FPL & Flexor Pollicis Longus & 1x \\
    APL & Abductor Pollicis Longus & 1x \\
    OP & Opponens Pollicis & 1x \\
    FDS & Flexor Digitorum Superficialis & 4x \\
    FDP & Flexor Digitorum Profundus & 4x \\
    EDC & Extensor Digitorum Communis & 4x \\
    EDM & Extensor digiti minimi & 1x \\
    RI & Radial Interosseous & 4x \\
    LU-RB & Lumbrical & 4x \\
    UI-UB & Palmar or Ulnar Interosseous & 4x \\
    \end{tabular}
    \end{center}
    \caption{List of muscles that are included in the MyoHand model.}
    \label{'Table:MusclesNames'}
\end{table}

\end{document}